\documentclass[11pt]{article}
\usepackage{fullpage}

\usepackage{hyperref}
\hypersetup{pdftex,colorlinks=true,allcolors=blue} 
\usepackage{hypcap} 

\usepackage[utf8]{inputenc} 
\usepackage[T1]{fontenc}    
\usepackage{url}            
\usepackage{booktabs}       
\usepackage{amsfonts}       
\usepackage{microtype}      

\usepackage{amsmath, amsthm, amssymb}
\usepackage{graphicx}
\usepackage{cite}
\usepackage{mathrsfs}
\usepackage{enumitem} 
\usepackage{lmodern}
\usepackage{comment}

\newtheorem{theorem}{Theorem}
\newtheorem{lemma}{Lemma}

\newtheorem{corollary}{Corollary}
\newtheorem{definition}{Definition}
\newcommand{\R}{\mathbb{R}}
\newcommand{\E}{\mathbb{E}}

\newcommand{\I}{\mathbf{1}}

\DeclareMathOperator*{\argmin}{arg\,min}

\def\sU{{\mathsf U}}
\def\sV{{\mathsf V}}
\def\sW{{\mathsf W}}
\def\sX{{\mathsf X}}
\def\sY{{\mathsf Y}}

\def\PP{{\mathbb P}}

\def\deq{\triangleq}
\def\wh#1{{\widehat{#1}}}

\newcommand\blfootnote[1]{%
	\begingroup
	\renewcommand\thefootnote{}\footnote{#1}%
	\addtocounter{footnote}{-1}%
	\endgroup
}

\title{Bayesian Learning for Dynamic Inference}
\author{Aolin Xu \and Peng Guan}
\date{}
\begin{document}

\maketitle

\begin{abstract}
	The traditional statistical inference is static, in the sense that the estimate of the quantity of interest does not affect the future evolution of the quantity. 
	In some sequential estimation problems however, the future values of the quantity to be estimated depend on the estimate of its current value. This type of estimation problems has been formulated as the dynamic inference problem. 
	In this work, we formulate the Bayesian learning problem for dynamic inference, where the unknown quantity-generation model is assumed to be randomly drawn according to a random model parameter. We derive the optimal Bayesian learning rules, both offline and online, to minimize the inference loss.
	Moreover, learning for dynamic inference can serve as a meta problem, such that all familiar machine learning problems, including supervised learning, imitation learning and reinforcement learning, can be cast as its special cases or variants.
	Gaining a good understanding of this unifying meta problem thus sheds light on a broad spectrum of machine learning problems as well.
\end{abstract}

\section{Introduction}
\subsection{Dynamic inference}
Traditional statistical estimation, or statistical inference in general is static, in the sense that the estimate of the quantity of interest does not affect the future evolution of the quantity. 
In some sequential estimation problems however, we do encounter the situation where the future value of the quantity to be estimated depends on the estimate of its current value. 
Examples include 1) stock price prediction by big investors, where the prediction of the tomorrow's price of a stock affects tomorrow's investment decision, which further changes the stock's supply-demand status and hence its price the day after tomorrow; 2) interactive product recommendation, where the estimate of a user's preference based on the user's activity leads to certain product recommendations to the user, which would in turn shape the user's future activity and preference; 3) behavior prediction in multi-agent systems, e.g.\ vehicles on the road, where the estimate of an adjacent vehicle's intention based on its current driving situation leads to a certain action of the ego vehicle, which can change the future driving situation and intention of the adjacent vehicle.
We may call such problems as \emph{dynamic inference}, which is formulated and studied in depth in \cite{dynamic_inf}. 
It is shown that the problem of dynamic inference can be converted to an Markov decision-making process (MDP), and the optimal estimation strategy can be derived through dynamic programming. 
We give a brief overview of the problem of dynamic inference in Section~\ref{sec:overview_di}.

\subsection{Learning for dynamic inference}\label{sec:intro_ldi}
There are two major ingredients in dynamic inference: the probability transition kernels of the quantity of interest given each observation, and the probability transition kernels of the next observation given the current observation and the estimate of the current quantity of interest. 
We may call them the \emph{quantity-generation model} and the \emph{observation-transition model}, respectively.
Solving the dynamic inference problem requires the knowledge of the two models.
However, in most of the practically interesting situations, we do not have such knowledge. Instead, we either have a training dataset from which we can learn these models or we can learn them on-the-fly during the inference.

In this work, we set up the learning problem in a \emph{Bayesian framework}, and derive the optimal learning rules, both offline (Section~\ref{sec:offline}) and online (Section~\ref{sec:online}), for dynamic inference under this framework.
Specifically, we assume the unknown models are elements in some parametric families of probability transition kernels, and the unknown model parameters are randomly drawn according to some prior distributions. 
The goal is then to find an optimal Bayesian learning rule, which can return an estimation strategy that minimizes the inference loss.
The approach we take toward this goal is converting the learning problem to an MDP with an augmented state, which consists of the current observation and a belief vector of the unknown parameters, and solving the MDP by dynamic programming over the augmented state space. 
The solution, though optimal, may still be computationally challenging unless the belief vector can be compactly represented. 
Nevertheless, it already has a greatly reduced search space compared to the original learning problem, and provides a theoretical basis for the design of more computationally efficient approximate solutions.

Perhaps equally importantly, the problem of learning for dynamic inference can serve as a meta problem, such that almost all familiar learning problems can be cast as its special cases or variants.
Examples include supervised learning, imitation learning, and reinforcement learning, including bandit and contextual bandit problems.
For instance, the Bayesian \textit{offline} learning for dynamic inference can be viewed as an extension of the \textit{behavior cloning} method in imitation learning \cite{Grimes, Englert, BC_from_obs}, in that it not only learns the demonstrator's action-generation model, but simultaneously learns a policy based on the learned model to minimize the overall imitation error. 
As another instance, the quantity to be estimated in dynamic inference may be viewed as a latent variable of the loss function, so that the Bayesian \textit{online} learning for dynamic inference can be viewed as \textit{Bayesian reinforcement learning} \cite{Feldbaum,Strens,Poupart,BRL_book}, where an optimal policy is learned by estimating the unknown loss function.
Learning for dynamic inference thus provides us with a unifying formulation of different learning problems. 
Having a good understanding of this problem is helpful for gaining better understandings of the other learning problems as well.

\subsection{Relation to existing works}
The problem of dynamic inference and learning for dynamic inference appear to be new, but it can be viewed from different angles, and is related to a variety of existing problems.
The most intimately related work is the original formulations of imitation learning \cite{Bagnell_il11}.
The online learning for dynamic inference is closely related to ans subsumes Bayesian reinforcement learning. Some recent study on Bayesian reinforcement learning and interactive decision making 
include \cite{stcomplx_idm21,ps_idm22}.

A problem formulation with a similar spirit in a minimax framework appear recently in \cite{online_dynamics}. In that work, an adversarial online learning problem where the action in each round affects the future observed data is set up. It may be viewed as \textit{adversarial online} learning for dynamic \textit{minimax} inference, from our standpoint. The advantage of the Bayesian formulation is that all the variables under consideration, including the unknown model parameters, are generated from some fixed joint distribution, thus the optimality of learning can be defined and the optimal learning rule can be derived. On the contrary, with the adversarial formulation, only certain definitions of regret can be studied.

The overall optimality proof technique we adopt is similar to those used in solving partially observed MDP (POMDP) and Bayesian reinforcement learning over the augmented belief space \cite{dual_control,Duff_thesis}.
Several proofs are adapted from the rigorous exposition of the optimality of the belief-state MDP reformulation of the POMDP \cite{MR_SOC_notes}.

As mentioned in the previous subsection, Bayesian learning for dynamic inference can be viewed as a unifying formulation for Bayesian imitation learning and Bayesian reinforcement learning. These problems are surveyed in \cite{IL_book,ChoiKim,Ramachandran} for relevant imitation learning, and in \cite{BRL_book,Dearden,Klenske,Guez,Michini} for relevant reinforcement learning.

\section{Overview of dynamic inference}\label{sec:overview_di}

\subsection{Problem formulation}
\begin{figure}[h]
	\centering
	\includegraphics[scale = 0.55]{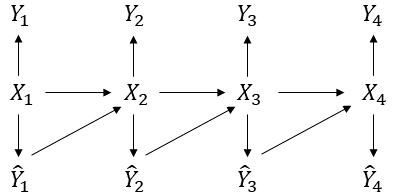}
	\caption{Bayesian network of the random variables under consideration with $n=4$. Here we assume the estimates are made with Markov estimators, such that $\wh Y_i = \psi_i(X_i)$.}
	\label{fig:BN_DI}
\end{figure}

The problem of an $n$-round dynamic inference is to estimate $n$ unknown quantities of interest $Y^n$ \emph{sequentially} based on observations $X^n$, where in the $i$th round of estimation, $X_i$ depends on the observation $X_{i-1}$ and the estimate $\wh Y_{i-1}$ of $Y_{i-1}$ in the previous round, while the quantity of interest $Y_i$ only depends on $X_i$, and the estimate $\wh Y_i$ of $Y_i$ can depend on everything available so far, namely $(X^{i},\wh Y^{i-1})$, through an estimator $\psi_i$ as $\wh Y_i = \psi_i(X^i,\wh Y^{i-1})$.
The sequence of estimators $\psi^n = (\psi_1,\ldots,\psi_n)$ constitute an \textit{estimation strategy}.
We assume to know the distribution $P_{X_1}$ of the initial observation, and the probability transition kernels $(K{\raisebox{-2pt}{$\scriptstyle X_i|X_{i-1}, \wh Y_{i-1}$}})_{i=2}^n$ and $(K_{Y_i|X_i})_{i=1}^n$.
These distributions and $\psi^n$ define a joint distribution of $(X^n,Y^n,\wh Y^n)$, all the variables under consideration.
The Bayesian network of the random variables in dynamic inference with a Markov estimation strategy, meaning that each estimator has the form $\psi_i:\sX\rightarrow\wh\sY$, is illustrated in Fig.~\ref{fig:BN_DI}.

The goal of dynamic inference can then be formally stated as finding an estimation strategy to minimize the accumulated expected loss over the $n$-rounds:
\begin{align}\label{eq:di_loss}
	\argmin_{\psi^n} \,\, \E\Big[ \sum_{i=1}^n \ell(X_i, Y_i, \wh Y_i) \Big] , \quad \wh Y_i = \psi_i(X^i,\wh Y^{i-1})
\end{align}
where $\ell:\sX\times\sY\times\wh\sY\rightarrow\R$ is a loss function that evaluates the estimate made in each round.
Compared with the traditional statistical inference under the Bayesian formulation, where the goal is to find an estimator $\psi$ of a random quantity $Y$ based on a jointly distributed observation $X$ to minimize $\E[\ell(Y,\psi(X))]$, we summarize the two distinctive features of dynamic inference in \eqref{eq:di_loss}:
\begin{itemize}[leftmargin=*]
	\item
	The joint distribution of the pair $(X_i,Y_i)$ changes in each round in a controlled manner, as it depends on $(X_{i-1},\wh Y_{i-1})$;
	\item
	The loss in each round is contextual, as it depends on $X_i$.
\end{itemize}

\subsection{Optimal estimation strategy for dynamic inference}
It is shown in \cite{dynamic_inf} that optimization problem in \eqref{eq:di_loss} is equivalent to
\begin{align}\label{eq:di_mdp}
	\argmin_{\psi^n} \,\, \E\Big[ \sum_{i=1}^n \bar \ell(X_i, \wh Y_i) \Big] ,
\end{align}
where 
$
	\bar \ell(x,\hat y) \deq \E[\ell(x,Y,\hat y)|X=x, \wh Y=\hat y] , 
$
and for any realization $(x_i,\hat y_i)$ of $(X_i,\wh Y_i)$, it can be computed as 
$
	\bar \ell(x_i,\hat y_i)=\E[\ell(x_i,Y_i,\hat y_i)|X_i=x_i]
$.
With this reformulation, the unknown quantities $Y_i$ do not appear in the loss function any more, and the optimization problem becomes a standard MDP.
The observations $X^n$ become the states in this MDP, the estimates $\wh Y^n$ become the actions, the probability transition kernel $K{\raisebox{-2pt}{$\scriptstyle X_i|X_{i-1}, \wh Y_{i-1}$}}$ now defines the controlled state transition, and any estimation strategy $\psi^n$ becomes a policy of this MDP. The goal becomes finding an optimal policy for this MDP to minimize the accumulated expected loss defined w.r.t.\ $\bar \ell$. The solution to the MDP will be an optimal estimation strategy for dynamic inference.

From the theory of MDP it is known that the optimal estimators $(\psi^*_1,\ldots,\psi^*_n)$ for the optimization problem in \eqref{eq:di_mdp} can be Markov, meaning that $\psi^*_i$ can take only $X_i$ as input, and the values of the optimal estimates $\psi^*_i(x)$ for $i=1,\ldots,n$ and $x\in\sX$ can be found via dynamic programming.
Define the functions $Q^*_i : \sX\times\wh Y\rightarrow\R$ and $V^*_i:\sX\rightarrow\R$ recursively as 
$
Q^*_n(x,\hat y) \deq \bar\ell (x, \hat y), 
$ 
$ 
V^*_i(x) \deq \min_{\hat y \in\wh \sY} Q^*_i (x, \hat y)  
$ 
for $  i=n,\ldots,1$, 
and
$
Q^*_i(x,\hat y) \deq \bar\ell (x, \hat y) + \E[V^*_{i+1}(X_{i+1})|X_i=x, \wh Y_i=\hat y] 
$ 
for $i=n-1,\ldots,1 $.
The optimal estimate to make in the $i$th round when $X_i=x$ is then
\begin{align}\label{eq:dp_policy}
\psi_i^*(x) \deq \argmin_{\hat y \in\wh \sY} Q^*_i (x, \hat y) .
\end{align}
It is shown that the estimators $(\psi^*_1,\ldots,\psi^*_n)$ defined in \eqref{eq:dp_policy} 
achieve the minimum in \eqref{eq:di_loss}.
	Moreover, For any $i=1,\ldots,n$ and any initial distribution $P_{X_i}$,
	\begin{align}
		\min_{\psi_i,\ldots,\psi_n} \,\, \E\Big[ \sum_{j=i}^n \ell(X_j, Y_j, \wh Y_j) \Big] = \E[V^*_i(X_i)] ,
	\end{align}
	with the minimum achieved by $(\psi^*_i,\ldots,\psi^*_n)$.
	As shown by the examples in \cite{dynamic_inf}, the implication of the optimal estimation strategy is that, in each round of estimation, the estimate to make is not necessarily the optimal single-round estimate in that round, but one which takes into account the accuracy in that round, and tries to steer the future observations toward those with which the quantities of interest tend to easy to estimate.

\section{Bayesian offline learning for dynamic inference}\label{sec:offline}
Solving dynamic inference requires the knowledge of the quantity-generation models $(K_{Y_i|X_i})_{i=1}^n$ and the observation-transition models $(K{\raisebox{-2pt}{$\scriptstyle X_i|X_{i-1}, \wh Y_{i-1}$}})_{i=2}^n$.
In most of the practically interesting situations however, we may not have such knowledge. Instead we may have a training dataset from which we can learn these models, or may learn them on-the-fly during inference.
In this section and the next one, we study the offline learning and the online learning problems for dynamic inference respectively, with \textit{unknown} quantity-generation models but \textit{known} observation transition models.
This is already a case of sufficient interest, as the observation-transition model in many problems, e.g. imitation learning, are available.
The proof techniques we develop carry over to the case where the observation-transition models are also unknown. In that case, the solution will have the same form, but a further-augmented state with a belief vector of the observation-transition model parameter; and the belief update has two parts, separately for the parameters of the quantity-generation model and the observation transition model.

\begin{figure}[t]
	\centering
	\includegraphics[scale = 0.55]{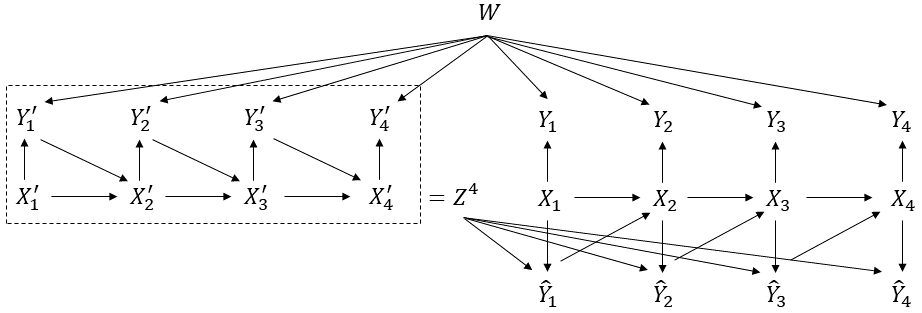}
	\caption{Bayesian network of the random variables in offline learning for dynamic inference with the imitation learning setup, with $m=n=4$. Here we assume the estimates are made with Markov estimators, such that $\wh Y_i = \psi_{m,i}(Z^m, X_i)$.}
	\label{fig:BN_LDI_IL}
\end{figure}

Formally, in this section we assume that the initial distribution $P_{X_1}$ and the probability transition kernels $(K{\raisebox{-2pt}{$\scriptstyle X_i|X_{i-1}, \wh Y_{i-1}$}})_{i=2}^n$ are still known, while the unknown $K_{Y_i|X_i}$'s are the same element $P_{Y|X,W}$ of a parametrized family of kernels $\{P_{Y|X,w}, w\in\sW\}$ and the unknown parameter $W$ is a random element of $\sW$ with prior distribution $P_W$.
The training data $Z^m$ consists of $m$ samples, and is drawn from some distribution $P_{Z^m|W}$ with $W$ as a parameter.
This setup is quite flexible, in that the $Z^m$ need not be generated in the same way as the data generated during inference.
One example is a setup similar to \textit{imitation learning}, where $Z^m = ((X'_1,Y'_1),\ldots,(X'_m,Y'_m))$ and
\begin{align}
P_{Z^m|W} = P_{X'_1}K_{Y'_1|X'_1} \prod_{i=2}^n K{\raisebox{-2pt}{$\scriptstyle X'_i|X'_{i-1}, Y'_{i-1}$}} K_{Y'_i|X'_i}
\end{align}
with $P_{X'_1}=P_{X_1}$, $(K{\raisebox{-2pt}{$\scriptstyle X'_i|X'_{i-1}, \wh Y'_{i-1}$}})_{i=2}^n = (K{\raisebox{-2pt}{$\scriptstyle X_i|X_{i-1}, \wh Y_{i-1}$}})_{i=2}^n$, and $K_{Y'_i|X'_i}=K_{Y_i|X_i}=P_{Y|X,W}$ for $i=1,\ldots,n$.
With a training dataset, we can define the \emph{offline-learned estimation strategy} for dynamic inference as follows.
\begin{definition}
	An offline-learned estimation strategy with an $m$-sample training dataset for an $n$-round dynamic inference is a sequence of estimators $\psi_m^n = (\psi_{m,1},\ldots,\psi_{m,n})$, where  $\psi_{m,i}:(\sX\times\wh\sY)^m\times\sX^{i}\times\wh\sY^{i-1}\rightarrow\wh\sY$ is the estimator for the $i$th round of estimation, which maps the dataset $Z^m$ as well as the past observations and estimates $(X^i,\wh Y^{i-1})$ up to the $i$th round to an estimate $\wh Y_i$ of $Y_i$, such that $\wh Y_i = \psi_{m,i}(Z^m,X^{i},\wh Y^{i-1})$, $i=1,\ldots,n$.
\end{definition}
Any specification of the above probabilistic models and an offline-learned estimation strategy determines a joint distribution of the random variables $(W,Z^m,X^n,Y^n,\wh Y^n)$ under consideration. The Bayesian network of the variables is shown in Fig.~\ref{fig:BN_LDI_IL}, where the training data is assumed to be generated in the imitation learning setup.
A crucial observation from the Bayesian network is that $W$ is conditionally independent of $(X^n,\wh Y^n)$ given $Z^m$, as the quantities of interest $Y^n$ are not observed.
In other words, given the training data, no more information about $W$ can be gained during inference.
We formally state this observation as the following lemma.
\begin{lemma}\label{lm:cond_ind_W_LDI}
	In offline learning for dynamic inference, the parameter $W$ is conditionally independent of the observations and the estimates $(X^n,\wh Y^n)$ during inference given the training data $Z^m$.
\end{lemma}

Given an offline-learned estimation strategy $\psi_m^{n}$ for an $n$-round dynamic inference with an $m$-sample training dataset, we can define its \emph{inference loss} as
$\E\big[ \sum_{i=1}^n \ell(X_i,Y_i,\wh Y_i) \big]$.
The goal of offline learning is to find an {offline-learned estimation strategy} to minimize the {inference loss}:
\begin{align}\label{eq:problem_LDI}
\argmin_{\psi_m^{n}} \E\Big[ \sum_{i=1}^n \ell(X_i,Y_i,\wh Y_i) \Big] , \quad \text{with $\wh Y_i = \psi_{m,i}(Z^m,X^{i},\wh Y^{i-1})$.}
\end{align}

\subsection{MDP reformulation}
\subsubsection{Equivalent expression of inference loss}
We first show that the inference loss in \eqref{eq:problem_LDI} can be expressed in terms of a loss function that does not take the unknown $Y_i$ as input.
\begin{theorem}\label{th:loss_LDI}
	For any offline-learned estimation strategy $\psi_m^{n}$, its inference loss can be written as
	\begin{align}\label{eq:loss_LDI}
	\E\Big[ \sum_{i=1}^n \ell(X_i,Y_i,\wh Y_i) \Big]  = \E\Big[ \sum_{i=1}^n \tilde{\ell}(\pi_m, X_i, \wh Y_i) \Big] ,
	\end{align}
	where $\pi_m(\cdot)\deq \PP[W\in\cdot|Z^m]$ is the posterior distribution of the kernel parameter $W$ given the training dataset $Z^m$, and  $\tilde{\ell}:\Delta\times\sX\times\wh\sY\rightarrow\R$, with $\Delta$ being the space of probability distributions on $\sW$, is defined as
	\begin{align}\label{eq:def_tilde_ell}
	\tilde{\ell}(\pi, x, \hat y) \deq \int_\sW \int_\sY \pi({\rm d}w) P_{Y|X,W}({\rm d}y | x, w) \ell(x, y, \hat y) .
	\end{align}
\end{theorem}
The proof is given in Appendix~\ref{appd:th:loss_LDI}.
Theorem~\ref{th:loss_LDI} states that the inference loss of an offline-learned estimation strategy $\psi_m^n$ is equal to 
\begin{align}\label{eq:def_inf_loss}
J(\psi_m^{n}) \deq \E\Big[ \sum_{i=1}^n \tilde\ell(\pi_m,X_i,\wh Y_i) \Big] ,
\end{align}
with $\wh Y_i = \psi_{m,i}(Z^m,X^{i},\wh Y^{i-1})$.
It follows that the offline learning problem in \eqref{eq:problem_LDI} can be equivalently written as
\begin{align}\label{eq:LDI_eq_loss}
\argmin_{\psi_m^n} J(\psi_m^n) .
\end{align}

\subsubsection{$(\pi_m,X_i)_{i=1}^n$ as a controlled Markov chain}
Next, we show that the sequence $(\pi_m,X_i)_{i=1}^n$ appearing in \eqref{eq:def_inf_loss} form a controlled Markov chain with $\wh Y^n$ as the control sequence.
In other words, the tuple $(\pi_m, X_{i+1})$ depends on the history $(\pi_m, X^i, \wh Y^i)$ only through $(\pi_m, X_i,  \wh Y_i)$, as formally stated in the following lemma.
\begin{lemma}\label{lm:controlled_MC_LDI}
	Given any offline-learned estimation strategy $\psi_m^n$, we have
	\begin{align}\label{eq:pf_controlled_MC_LDI}
	\PP\big[ (\pi_m, X_{i+1}) \in A\times B \big| \pi_m, X^i, \wh Y^i \big ] =  \I\{\pi_m \in A\} \PP\big[ X_{i+1} \in B | X_i, \wh Y_i \big] 
	\end{align}
	for any Borel sets $A\subset\Delta$ and $B\subset\sX$, any realization of $(\pi_m, X^i, \wh Y^i)$, and any $i=1,\ldots,n-1$.
\end{lemma}
The proof is given in Appendix~\ref{appd:lm:controlled_MC_LDI}.


\subsubsection{Optimality of Markov offline-learned estimators}
Furthermore, the next three lemmas will show that the search space of the minimization problem in \eqref{eq:LDI_eq_loss}
can be restricted to Markov offline-learned estimators $\bar\psi_{m,i}: \Delta\times\sX\rightarrow\sY$, such that $\wh Y_i = \bar\psi_{m,i}(\pi_m, X_i)$.
We start with a generalization of Blackwell's principle of irrelevant information.
\begin{lemma}[Generalized Blackwell's principle of irrelevant information]\label{lm:gen_Blackwell}
	For any fixed functions $\ell:\sY\times\wh\sY\rightarrow \R$ and $f:\sX\rightarrow\sY$, the following equality holds:
	\begin{align}\label{eq:gen_Blackwell}
	\min_{g:\sX\rightarrow\wh\sY} \E\big[\ell\big(f(X),g(X)\big)\big] = \min_{g:\sY\rightarrow\wh\sY} \E\big[\ell\big(f(X), g(f(X))\big)\big] .
	\end{align}
\end{lemma}
\noindent{\bf Remark.}
The original Blackwell's principle of irrelevant information, stating that for any fixed function $\ell:\sY\times\wh\sY\rightarrow \R$,
\begin{align}
\min_{g:\sX\times\sY\rightarrow\wh\sY} \E\big[\ell\big(Y,g(X,Y)\big)\big] = \min_{g:\sY\rightarrow\wh\sY} \E\big[\ell\big(Y, g(Y)\big)\big] ,
\end{align}
can be seen as a special case of the above lemma.

The proof of Lemma~\ref{lm:gen_Blackwell} is given in Appendix~\ref{appd:lm:gen_Blackwell}.
The first application of Lemma~\ref{lm:gen_Blackwell} is to prove that the last estimator of an optimal offline-learned estimation strategy can be replaced by a Markov one, which preserves the optimality.
\begin{lemma}[Last-round lemma for offline learning]\label{lm:last_round}
	Given any offline-learned estimation strategy $\psi_m^n$, there exists a Markov offline-learned estimator $\bar\psi_{m,n}:\Delta\times\sX\rightarrow\wh\sY$, such that
	\begin{align}
	J(\psi_{m,1},\ldots,\psi_{m,n-1}, \bar\psi_{m,n}) \le J(\psi_m^{n}) .
	\end{align}
\end{lemma}
The proof is given in Appendix~\ref{appd:lm:last_round}.
Lemma~\ref{lm:gen_Blackwell} can be further used to prove that whenever the last offline-learned estimator is Markov, the preceding estimator can also be replaced by a Markov one which preserves the optimality.
\begin{lemma}[$(i-1)$th-round lemma for offline learning]\label{lm:i-1th_round}
	For any $i\ge 2$, given any offline-learned estimation strategy $(\psi_{m,1},\ldots,\psi_{m,i-1}, \bar\psi_{m,i})$ for an $i$-round dynamic inference with an $m$-sample training dataset, if the offline-learned estimator for the $i$th round of estimation is a Markov one $\bar \psi_{m,i}:\Delta\times\sX\rightarrow\wh\sY$, then there exists a Markov offline-learned estimator $\bar\psi_{m,i-1}:\Delta\times\sX\rightarrow\wh\sY$ for the $(i-1)$th round, such that
	\begin{align}
	J(\psi_{m,1},\ldots,\psi_{m,i-2}, \bar\psi_{m,i-1}, \bar\psi_{m,i}) \le J(\psi_{m,1},\ldots,\psi_{m,i-1}, \bar\psi_{m,i}) .
	\end{align}
\end{lemma}
The proof is given in Appendix~\ref{appd:lm:i-1th_round}.
With Lemma~\ref{lm:last_round} and Lemma~\ref{lm:i-1th_round}, we can prove the optimality of Markov offline-learned estimators, as given in Appendix~\ref{appd:th:Markov_LDI}.
\begin{theorem}\label{th:Markov_LDI}
	The minimum of $J(\psi_m^n)$ in \eqref{eq:LDI_eq_loss}
	can be achieved by an offline-learned estimation strategy $\bar\psi_m^n$ with Markov estimators $\bar\psi_{m,i}:\Delta\times\sX\rightarrow\wh\sY$, $i=1,\ldots,n$, such that $\wh Y_i = \bar\psi_{m,i}(\pi_m,X_i)$.
\end{theorem}

\subsubsection{Conversion to MDP}
Theorem~\ref{th:loss_LDI} and Theorem~\ref{th:Markov_LDI} with Lemma~\ref{lm:controlled_MC_LDI} imply that the original offline learning problem in \eqref{eq:problem_LDI} is equivalent to
\begin{align}\label{eq:LDT_MDP}
\argmin_{ \psi_m^n} \E\Big[ \sum_{i=1}^n \tilde{\ell}(\pi_m, X_i, \wh Y_i) \Big] , \quad\wh Y_i = \psi_{m,i}(\pi_m, X_i) ,
\end{align}
and the sequence $(\pi_m,  X_i)_{i=1}^n$ is a controlled Markov chain driven by $\wh Y^n$.
With this reformulation, we see that the offline learning problem becomes a standard MDP.
The tuples $(\pi_m,X_i)_{i=1}^n$ become the states in this MDP, the estimates $\wh Y^n$ become the actions, the probability transition kernel $P{\raisebox{-2pt}{$\scriptstyle (\pi_m,X_i)|(\pi_m,X_{i-1}), \wh Y_{i-1}$}}$ now defines the controlled state transition, and any Markov offline-learned estimation strategy $\psi_m^n$ becomes a policy of this MDP. The goal of learning becomes finding the optimal policy of the MDP to minimize the accumulated expected loss defined w.r.t.\ $\tilde \ell$. The solution to this MDP will be an optimal offline-learned estimation strategy for dynamic inference.

\subsection{Solution via dynamic programming}
\subsubsection{Optimal offline-learned estimation strategy}
From the theory of MDP it is known that the optimal policy for the MDP in \eqref{eq:LDT_MDP}, namely the optimal offline-learned estimation strategy, can be found via dynamic programming.
To derive the optimal estimators, define the functions $Q^*_{m,i} : \Delta\times\sX\times\wh Y\rightarrow\R$ and $V^*_{m,i}:\Delta\times\sX\rightarrow\R$ for offline learning recursively for $i=n,\ldots,1$ as 
$
Q^*_{m,n}(\pi, x,\hat y) \deq \tilde\ell (\pi, x, \hat y) ,
$ and
\begin{align}
V^*_{m,i}(\pi, x) &\deq \min_{\hat y \in\wh \sY} Q^*_{m,i} (\pi, x, \hat y), \quad i = n,\ldots, 1   \label{eq:dp_LDI_V} \\
Q^*_{m,i}(\pi,x,\hat y) &\deq \tilde\ell (\pi, x, \hat y) + \E[V^*_{m,i+1}(\pi, X_{i+1}) | X_i=x, \wh Y_i=\hat y], \quad i = n-1,\ldots,1 \label{eq:dp_LDI_Q} 
\end{align}
with $\tilde{\ell}$ is as defined in \eqref{eq:def_tilde_ell}, and the conditional expectation in \eqref{eq:dp_LDI_Q} is taken w.r.t.\ $X_{i+1}$.
The optimal offline-learned estimate to make in the $i$th round when $\pi_m=\pi$ and $X_i=x$ is then
\begin{align}\label{eq:dp_LDI_policy}
\psi_{m,i}^*(\pi, x) \deq \argmin_{\hat y \in\wh \sY} Q^*_{m,i} (\pi, x, \hat y) .
\end{align}

\subsubsection{Minimum inference loss and loss-to-go}
For any offline-learned estimation strategy $\psi_m^n$, we can define its loss-to-go in the $i$th round of estimation when $\pi_m=\pi$ and $X_i=x$ as
\begin{align}\label{eq:def_Vmi}
V_{m,i}(\pi, x; \psi_m^n) \deq \E\Big[ \sum_{j=i}^n \ell(X_j, Y_j, \wh Y_j) \Big| \pi_m = \pi, X_i=x\Big] ,
\end{align}
which is the conditional expected loss accumulated from the $i$th round to the final round when $(\psi_{m,i},\ldots,\psi_{m,n})$ are used as the offline-learned estimators, given that the posterior distribution of the kernel parameter $W$ given the training dataset $Z^m$ is $\pi$ and the observation in the $i$th round is $x$.
The following theorem states that the offline-learned estimation strategy $(\psi^*_{m,1},\ldots,\psi^*_{m,n})$ derived from dynamic programming not only achieves the minimum inference loss over the $n$ rounds, but also achieves the minimum loss-to-go in each round with any training dataset and any observation in that round.
\begin{theorem}\label{th:opt_dp_LDI}
	The offline-learned estimators $(\psi^*_{m,1},\ldots,\psi^*_{m,n})$ defined in \eqref{eq:dp_LDI_policy} according to the recursion in \eqref{eq:dp_LDI_V} and \eqref{eq:dp_LDI_Q} constitute an optimal offline-learned estimation strategy for dynamic inference, which achieves the minimum in \eqref{eq:problem_LDI}.
	Moreover, for any Markov offline-learned estimation strategy $\psi_m^n$, with $\psi_{m,i}:\Delta\times\sX\rightarrow\sY$, its loss-to-go satisfies
	\begin{align}
	V_{m,i}(\pi, x; \psi_m^n) \ge V^*_{m,i}(\pi,x) \label{eq:opt_V_dp_LDI}
	\end{align}
	for all $\pi\in\Delta$, $x\in\sX$ and $i=1,\ldots,n$,
	where the equality holds if $\psi_{m,j}(\pi,x)=\psi_{m,j}^*(\pi,x)$ for all $\pi\in\Delta$, $x\in\sX$ and $j\ge i$.
\end{theorem}
The proof is given in Appendix~\ref{appd:th:opt_dp_LDI}.
A consequence of Theorem~\ref{th:opt_dp_LDI} is that in offline learning for dynamic inference, the minimum expected loss accumulated from the $i$th round to the final round can be expressed in terms of $V^*_{m,i}$, as stated in the following corollary.
\begin{corollary}
	In offline learning for dynamic inference, for any $i$ and any initial distribution $P_{X_i}$,
	\begin{align}
	\min_{\psi_{m,i},\ldots,\psi_{m,n}} \,\, \E\Big[ \sum_{j=i}^n \ell(X_j, Y_j, \wh Y_j) \Big] = \E[V^*_{m,i}(\pi_m, X_i)] ,
	\end{align}
	and the minimum is achieved by the estimators $(\psi^*_{m,i},\ldots,\psi^*_{m,n})$ defined in \eqref{eq:dp_LDI_policy}.
\end{corollary}

\section{Bayesian online learning for dynamic inference}\label{sec:online}
In the setup  of offline learning for dynamic inference, we assume that before the inference takes place, a training dataset $Z^m$ drawn from some distribution $P_{Z^m|W}$ is observed, and $W$ can be estimated from $Z^m$.
In the online learning setup, we assume that there is no training dataset available before the inference; instead, during the inference, after an estimate $\wh Y_i$ is made in each round, the true value $Y_i$ is revealed, and $W$ can be estimated on-the-fly in each round from all the observations available so far.

\begin{figure}[h]
	\centering
	\includegraphics[scale = 0.55]{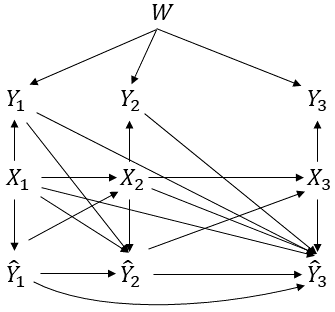}
	\caption{Bayesian network of variables in online learning for dynamic inference, with $n=3$. 
	}
	\label{fig:BN_onLDI_IL}
\end{figure}

Same as the offline learning setup, we assume that during inference, the initial distribution $P_{X_1}$ and the probability transition kernels $K{\raisebox{-2pt}{$\scriptstyle X_i|X_{i-1}, \wh Y_{i-1}$}}$, $i=1,\ldots,n$ are still known, while the unknown $K_{Y_i|X_i}$'s are the same element $P_{Y|X,W}$ of a parametrized family of kernels $\{P_{Y|X,w}, w\in\sW\}$ and the unknown kernel parameter $W$ is a random element of $\sW$ with prior distribution $P_W$.
We can define the \emph{online-learned estimation strategy} for dynamic inference as follows. 
Note that we overload the notations $\psi_i$ as an online-learned estimator and $Z_i$ as $(X_i,Y_i)$ throughout this section.
\begin{definition}
	An online-learned estimation strategy for an $n$-round dynamic inference is a sequence of estimators $\psi^n = (\psi_{1},\ldots,\psi_{n})$, where  $\psi_{i}:(\sX\times\sY)^{i-1}\times\wh\sY^{i-1}\times\sX \rightarrow \wh\sY$ is the estimator in the $i$th round of estimation, which maps the past observations $Z^{i-1}= (X_j,Y_j)_{j=1}^{i-1}$ and estimates $\wh Y^{i-1}$ in addition to a new observation $X_i$ to an estimate $\wh Y_i$ of $Y_i$, such that $\wh Y_i = \psi_i(Z^{i-1},\wh Y^{i-1}, X_i)$.
\end{definition}
The Bayesian network of all the random variables $(W,X^n,Y^n,\wh Y^n)$ in online learning for dynamic inference is shown in Fig.~\ref{fig:BN_onLDI_IL}.
A crucial observation from the Bayesian network is that $W$ is conditionally independent of $(X_i,\wh Y^{i})$ given $Z^{i-1}$, as stated in the following lemma.
\begin{lemma}\label{lm:cond_ind_W_onLDI}
	In online learning for dynamic inference, in the $i$th round of estimation, the kernel parameter $W$ is conditionally independent of the current observation $X_i$ and the estimates $\wh Y^{i}$ up to the $i$th round given the past observations $Z^{i-1}$.
\end{lemma}

Same as the offline learning setup, given an online-learned estimation strategy $\psi^{n}$, we can define its {inference loss} as
$\E\big[ \sum_{i=1}^n \ell(X_i,Y_i,\wh Y_i) \big]$.
The goal of online learning for an $n$-round dynamic inference is to find an {online-learned estimation strategy} to minimize the {inference loss}:
\begin{align}\label{eq:problem_onLDI}
\argmin_{\psi^{n}} \E\Big[ \sum_{i=1}^n \ell(X_i,Y_i,\wh Y_i) \Big] , \quad\text{with $\wh Y_i = \psi_i(Z^{i-1},\wh Y^{i-1}, X_i)$.}
\end{align}

\subsection{MDP reformulation}
\subsubsection{Equivalent expression of inference loss}
We first show that the inference loss in \eqref{eq:problem_onLDI} can be expressed in terms of a loss function that does not take the unknown $Y_i$ as input.
\begin{theorem}\label{th:loss_onLDI}
	For any online-learned estimation strategy $\psi^{n}$, its inference loss can be written as
	\begin{align}\label{eq:loss_onLDI}
	\E\Big[ \sum_{i=1}^n \ell(X_i,Y_i,\wh Y_i) \Big]  = \E\Big[ \sum_{i=1}^n \tilde{\ell}(\pi_i, X_i, \wh Y_i) \Big] ,
	\end{align}
	where $\pi_i(\cdot)\deq \PP[W\in\cdot|Z^{i-1}]$ is the posterior distribution of the kernel parameter $W$ given the past observations $Z^{i-1}$  to the $i$th round, and $\tilde{\ell}:\Delta\times\sX\times\wh\sY\rightarrow\R$, with $\Delta$ being the space of probability distributions on $\sW$, is defined in the same way as in \eqref{eq:def_tilde_ell},
	\begin{align}
	\tilde{\ell}(\pi, x, \hat y) = \int_\sW \int_\sY \pi({\rm d}w) P_{Y|X,W}({\rm d}y | x, w) \ell(x, y, \hat y) .
	\end{align}
\end{theorem}
The proof is given in Appendix~\ref{appd:th:loss_onLDI}.
Theorem~\ref{th:loss_onLDI} states that the inference loss of an online-learned estimation strategy $\psi^n$ is equal to 
\begin{align}\label{eq:def_inf_loss_on}
J(\psi^{n}) = \E\Big[ \sum_{i=1}^n \tilde\ell(\pi_i,X_i,\wh Y_i) \Big] , \quad \text{with $\wh Y_i = \psi_i(Z^{i-1},\wh Y^{i-1}, X_i)$.}
\end{align}
It follows that the learning problem in \eqref{eq:problem_onLDI} can be equivalently written as
\begin{align}\label{eq:onLDI_eq_loss}
\argmin_{\psi^n} J(\psi^n) .
\end{align}

\subsubsection{$(\pi_i,X_i)_{i=1}^n$ as a controlled Markov chain}
Next, we show that the sequence $(\pi_i,X_i)_{i=1}^n$ appearing in \eqref{eq:def_inf_loss_on} form a controlled Markov chain with $\wh Y^n$ as the control sequence.
In other words, the tuple $(\pi_{i+1}, X_{i+1})$ depends on the history $(\pi^i, X^i, \wh Y^i)$ only through $(\pi_i, X_i,  \wh Y_i)$, as formally stated in the following lemma.
\begin{lemma}\label{lm:controlled_MC_onLDI}
	There exists a function $f:\Delta\times\sX\times\sY \rightarrow \Delta$, such that given any learned estimation strategy $\psi^n$, we have
	\begin{align}\label{eq:pf_controlled_MC_onLDI}
	\PP\big[ & (\pi_{i+1}, X_{i+1}) \in A\times B \big| \pi^i, X^i, \wh Y^i \big ] = \nonumber \\ 
	&  \int_\sW \int_\sY \pi_i({\rm d}w) P_{Y|X,W}({\rm d}y_i | X_i, w)  \PP[f(\pi_{i}, X_i, y_i) \in A] \PP\big[ X_{i+1} \in B | X_i, \wh Y_i \big] 
	\end{align}
	for any Borel sets $A\subset\Delta$ and $B\subset\sX$, any realization of $(\pi_i, X^i, \wh Y^i)$, and any $i=1,\ldots,n-1$.
\end{lemma}
Lemma~\ref{lm:controlled_MC_onLDI} is proved in Appendix~\ref{appd:lm:controlled_MC_onLDI}, based on the auxiliary lemma below proved in Appendix~\ref{appd:lm:cond_P}.
\begin{lemma}\label{lm:cond_P}
	For a generic random tuple $(T,U,V)\in \mathsf T\times\sU\times\sV$ that forms a Markov chain $T-U-V$, we have
	\begin{align}
	\PP\big[ V \in  A \big| P_{V|U}(\cdot|U) = p, T \in  B \big] = p( A)
	\end{align}
	for any Borel sets $ A\in \mathsf V$ and $ B\in\mathsf T$, and any probability distribution $p$ on $\sV$.
\end{lemma}

\subsubsection{Optimality of Markov online-learned estimators}
The next two lemmas will show that the search space of the minimization problem in \eqref{eq:onLDI_eq_loss}
can be restricted to Markov online-learned estimators $\bar\psi_i: \Delta\times\sX\rightarrow\sY$, such that $\wh Y_i = \bar\psi_i(\pi_i, X_i)$.
In parallel to the discussion of the offline learning, we first prove that the last estimator of an optimal online-learned estimation strategy can be replaced by a Markov one, which preserves the optimality.
\begin{lemma}[Last-round lemma for online learning]\label{lm:last_round_on}
	Given any online-learned estimation strategy $\psi^n$, there exists a Markov online-learned estimator $\bar\psi_{n}:\Delta\times\sX\rightarrow\wh\sY$, such that
	\begin{align}
	J(\psi_{1},\ldots,\psi_{n-1}, \bar\psi_{n}) \le J(\psi^{n}) .
	\end{align}
\end{lemma}
The proof is given in Appendix~\ref{appd:lm:last_round_on}.
We further prove that whenever the last online-learned estimator is Markov, the preceding estimator can be replaced by a Markov one which preserves the optimality.
\begin{lemma}[$(i-1)$th-round lemma for online learning]\label{lm:i-1th_round_on}
	For any $i\ge 2$, given any online-learned estimation strategy $(\psi_{1},\ldots,\psi_{i-1}, \bar\psi_{i})$ for an $i$-round dynamic inference, if the last estimator is a Markov one $\bar \psi_{i}:\Delta\times\sX\rightarrow\wh\sY$, then there exists a Markov onlined-learned estimator $\bar\psi_{i-1}:\Delta\times\sX\rightarrow\wh\sY$ for the $(i-1)$th round, such that
	\begin{align}
	J(\psi_{1},\ldots,\psi_{i-2}, \bar\psi_{i-1}, \bar\psi_{i}) \le J(\psi_{1},\ldots,\psi_{i-1}, \bar\psi_{i}) .
	\end{align}
\end{lemma}
The proof is given in Appendix~\ref{appd:lm:i-1th_round_on}.
With Lemma~\ref{lm:last_round} and Lemma~\ref{lm:i-1th_round}, we can prove the optimality of Markov online-learned estimators, as given in Appendix~\ref{appd:th:Markov_onLDI}.
\begin{theorem}\label{th:Markov_onLDI}
	The minimum of $J(\psi^n)$ in \eqref{eq:onLDI_eq_loss}
	can be achieved by a online-learned estimation strategy $\bar\psi^n$ with Markov online-learned estimators $\bar\psi_{i}:\Delta\times\sX\rightarrow\wh\sY$, such that $\wh Y_i = \bar\psi_{i}(\pi_i,X_i)$.
\end{theorem}

\subsubsection{Conversion to MDP}
Theorem~\ref{th:loss_onLDI} and Theorem~\ref{th:Markov_onLDI} with Lemma~\ref{lm:controlled_MC_onLDI} imply that the original online learning problem in \eqref{eq:problem_onLDI} is equivalent to
\begin{align}\label{eq:onLDT_MDP}
\argmin_{ \psi^n} \E\Big[ \sum_{i=1}^n \tilde{\ell}(\pi_i, X_i, \wh Y_i) \Big] , \quad \wh Y_i = \psi_{i}(\pi_i, X_i)
\end{align}
and the sequence $(\pi_i,X_i)_{i=1}^n$ is a controlled Markov chain driven by $\wh Y^n$.
With this reformulation, we see that the online learning problem becomes a standard MDP.
The tuples $(\pi_i,X_i)_{i=1}^n$ become the states in this MDP, the estimates $\wh Y^n$ become the actions, the probability transition kernel $P{\raisebox{-2pt}{$\scriptstyle (\pi_i,X_i)|(\pi_{i-1},X_{i-1}), \wh Y_{i-1}$}}$ now defines the controlled state transition, and any Markov online-learned estimation strategy $\psi^n$ becomes a policy of this MDP. The goal of online learning becomes finding the optimal policy of the MDP to minimize the accumulated expected loss defined w.r.t.\ $\tilde \ell$. The solution to this MDP will be an optimal online-learned estimation strategy for dynamic inference.

\subsection{Solution via dynamic programming}
\subsubsection{Optimal online-learned estimation strategy}
From the theory of MDP it is known that the optimal policy for the MDP in \eqref{eq:onLDT_MDP}, namely the optimal online-learned estimation strategy, can be found via dynamic programming.
To derive the optimal estimators, define the functions $Q^*_{i} : \Delta\times\sX\times\wh Y\rightarrow\R$ and $V^*_{i}:\Delta\times\sX\rightarrow\R$ for online learning recursively for $i=n,\ldots,1$ as
$
Q^*_{n}(\pi, x,\hat y) \deq \tilde\ell (\pi, x, \hat y) ,
$ and
\begin{align}
V^*_{i}(\pi, x) &\deq \min_{\hat y \in\wh \sY} Q^*_{i} (\pi, x, \hat y), \quad i = n,\ldots, 1   \label{eq:dp_onLDI_V} \\
Q^*_{i}(\pi,x,\hat y) &\deq \tilde\ell (\pi, x, \hat y) + \E[V^*_{i+1}(\pi_{i+1}, X_{i+1}) | \pi_i = \pi, X_i=x, \wh Y_i=\hat y], \, i = n-1,\ldots,1 \label{eq:dp_onLDI_Q} 
\end{align}
with $\tilde{\ell}$ is as defined in \eqref{eq:def_tilde_ell}, and the conditional expectation in \eqref{eq:dp_onLDI_Q} is taken w.r.t.\ $(\pi_{i+1}, X_{i+1})$.
The optimal online-learned estimate to make in the $i$th round when $\pi_i=\pi$ and $X_i=x$ is then
\begin{align}\label{eq:dp_onLDI_policy}
\psi_{i}^*(\pi, x) \deq \argmin_{\hat y \in\wh \sY} Q^*_{i} (\pi, x, \hat y) .
\end{align}

\subsubsection{Minimum inference loss and loss-to-go}
For any online-learned estimation strategy $\psi^n$, we can define its loss-to-go in the $i$th round of estimation when $\pi_i=\pi$ and $X_i=x$ as
\begin{align}\label{eq:def_Vi_on}
V_{i}(\pi, x; \psi^n) \deq \E\Big[ \sum_{j=i}^n \ell(X_j, Y_j, \wh Y_j) \Big| \pi_i = \pi, X_i=x\Big] ,
\end{align}
which is the conditional expected loss accumulated from the $i$th round to the final round when $(\psi_{i},\ldots,\psi_{n})$ are used as the learned estimators, given that in the $i$th round the posterior distribution of the kernel parameter $W$ given the past observations $Z^{i-1}$ is $\pi$ and the observation $X_i$ is $x$.
The following theorem states that the online-learned estimation strategy $(\psi^*_{1},\ldots,\psi^*_{n})$ derived from dynamic programming not only achieves the minimum inference loss over the $n$ rounds, but also achieves the minimum loss-to-go in each round with any past and current observations in that round.
\begin{theorem}\label{th:opt_dp_onLDI}
	The online-learned estimators $(\psi^*_{1},\ldots,\psi^*_{n})$ defined in \eqref{eq:dp_onLDI_policy} according to the recursion in \eqref{eq:dp_onLDI_V} and \eqref{eq:dp_onLDI_Q} constitute an optimal online-learned estimation strategy for dynamic inference, which achieves the minimum in \eqref{eq:problem_onLDI}.
	Moreover, for any Markov online-learned estimation strategy $\psi^n$, with $\psi_{i}:\Delta\times\sX\rightarrow\sY$, its loss-to-go satisfies
	\begin{align}
	V_{i}(\pi, x; \psi^n) \ge V^*_{i}(\pi,x) \label{eq:opt_V_dp_onLDI}
	\end{align}
	for all $\pi\in\Delta$, $x\in\sX$ and $i=1,\ldots,n$,
	where the equality holds if $\psi_{j}(\pi,x)=\psi_{j}^*(\pi,x)$ for all $\pi\in\Delta$, $x\in\sX$ and $j\ge i$.
\end{theorem}
The proof is given in Appendix~\ref{appd:th:opt_dp_onLDI}.
A consequence of Theorem~\ref{th:opt_dp_onLDI} is that in online learning for dynamic inference, the minimum expected loss accumulated from the $i$th round to the final round can be expressed in terms of $V^*_{i}$, as stated in the following corollary.
\begin{corollary}
	In online learning for dynamic inference, for any $i$ and any initial distribution $P_{X_i}$,
	\begin{align}
	\min_{\psi_{i},\ldots,\psi_{n}} \,\, \E\Big[ \sum_{j=i}^n \ell(X_j, Y_j, \wh Y_j) \Big] = \E[V^*_{i}(\pi_i, X_i)] ,
	\end{align}
	and the minimum is achieved by the estimators $(\psi^*_{i},\ldots,\psi^*_{n})$ defined in \eqref{eq:dp_onLDI_policy}.
\end{corollary}



\appendix

\section{Proof of Theorem~\ref{th:loss_LDI}}\label{appd:th:loss_LDI}
	For each $i=1,\ldots,n$, we have
	\begin{align}
		&\E\big[ {\ell}( X_i, Y_i, \wh Y_i) \big| Z^m,  X^i, \wh Y^{i-1} \big] \nonumber \\
		=& \int_\sY  P_{Y_i|Z^m,  X^i, \wh Y^{i-1}}({\rm d}y) \ell(X_i, y, \wh Y_i) \label{eq:pf_LDI_loss_1} \\
		=& \int_\sW \int_\sY P_{W|Z^m,  X^i, \wh Y^{i-1}}({\rm d}w) P_{Y_i|Z^m,  X^i, \wh Y^{i-1},W=w}({\rm d}y) \ell(X_i, y, \wh Y_i) \\
		=& \int_\sW \int_\sY \pi_m({\rm d}w) P_{Y|X,W}({\rm d}y | X_i, w) \ell(X_i, y, \wh Y_i) \label{eq:pf_LDI_loss_2} \\
		=&\tilde{\ell}(\pi_m, X_i, \wh Y_i) ,
	\end{align}
	where \eqref{eq:pf_LDI_loss_1} is due to the fact that $X_i$ and $\wh Y_i$ are determined by $(Z^m, X^i, \wh Y^{i-1})$; and \eqref{eq:pf_LDI_loss_2} follows from the fact that $W$ is conditionally independent of $(X^i, \wh Y^{i-1})$ given $Z^m$ as stated in Lemma~\ref{lm:cond_ind_W_LDI}, and the fact that $Y_i$ is conditionally independent of $(Z^m, X^{i-1}, \wh Y^{i-1})$ given $(X_i,W)$.
	With the above equality and the fact that 
	\begin{align}
		\E\Big[ \sum_{i=1}^n \ell(X_i,Y_i,\wh Y_i) \Big] =  \sum_{i=1}^n \E\big[ \E[ {\ell}( X_i, Y_i, \wh Y_i) | Z^m,  X^i, \wh Y^{i-1} ] \big] ,
	\end{align}
	we obtain \eqref{eq:loss_LDI}.

\section{Proof of Lemma~\ref{lm:controlled_MC_LDI}}\label{appd:lm:controlled_MC_LDI}
	For any offline-learned estimation strategy $\psi_m^n$, any Borel sets $A\subset\Delta$ and $B\subset\sX$, and any realization of $(\pi_m, X^i, \wh Y^i)$,
	\begin{align}
	\PP\big[ (\pi_m, X_{i+1}) \in A\times B \big| \pi_m, X^i, \wh Y^i \big ] 
	&= \PP\big[ \pi_m \in A \big| \pi_m \big] \PP\big[ X_{i+1} \in B | \pi_m, X^i, \wh Y^i \big] \\
	&= \I\{\pi_m \in A\} \PP\big[ X_{i+1} \in B | X_i, \wh Y_i \big] 
	\end{align}
	where the second equality is due to the fact that $X_{i+1}$ is conditionally independent of $(\pi_m, X^{i-1}, \wh Y^{i-1})$ given $(X_i,\wh Y_i)$.
	This proves the claim, and we can see that the right side of \eqref{eq:pf_controlled_MC_LDI} only depends on $(\pi_m, X_i,  \wh Y_i)$.

\section{Proof of Lemma~\ref{lm:gen_Blackwell}}\label{appd:lm:gen_Blackwell}.
	The left side of \eqref{eq:gen_Blackwell} is the Bayes risk of estimating $f(X)$ based on $X$, defined w.r.t.\ the loss function $\ell$, which can be written as $R_\ell(f(X)|X)$;  while the right side of \eqref{eq:gen_Blackwell} is the Bayes risk of estimating $f(X)$ based on $f(X)$ itself, also defined w.r.t.\ the loss function $\ell$, which can be written as $R_\ell(f(X)|f(X))$.
	It follows from a data processing inequality of the generalized conditional entropy 
	that
	\begin{align}
	R_\ell(f(X)|X) \le R_\ell(f(X)|f(X)) ,
	\end{align}
	as $f(X)-X-f(X)$ form a Markov chain.
	If follows from the same data processing inequality that
	\begin{align}
	R_\ell(f(X)|X) \ge R_\ell(f(X)|f(X)) ,
	\end{align}
	as $X-f(X)-f(X)$ also form a Markov chain. Hence $R_\ell(f(X)|X)=R_\ell(f(X)|f(X))$, which proves the claim.

\section{Proof of Lemma~\ref{lm:last_round}}\label{appd:lm:last_round}
	The inference loss of $\psi_m^n$ can be written as
	\begin{align}
		J(\psi_m^{n}) 
		&= \E\Big[\sum_{i=1}^{n-1} \tilde{\ell}\big((\pi_m, X_i), \wh Y_i\big)\Big]
		+ \E\big[ \tilde{\ell}\big((\pi_m, X_n), \psi_{m,n}(Z^m, X^n, \wh Y^{n-1}) \big)\big] . \label{eq:pf_last_round_lm_2}
	\end{align}
	Since the first expectation in \eqref{eq:pf_last_round_lm_2} does not depend on $\psi_{m,n}$, it suffices to show that there exists a learned estimator $\bar\psi_{m,n}:\Delta\times\sX\rightarrow\wh\sY$, such that 
	\begin{align}
		\E\big[ \tilde{\ell}\big((\pi_m, X_n), \bar\psi_{m,n}(\pi_m, X_n) \big)\big]  \le
		\E\big[ \tilde{\ell}\big((\pi_m, X_n), \psi_{m,n}(Z^m, X^n, \wh Y^{n-1}) \big)\big] .
	\end{align}
	The existence of such an estimator is guaranteed by Lemma~\ref{lm:gen_Blackwell}, as $(\pi_m, X_n)$ is a function of $(Z^m, X^n, \wh Y^{n-1})$.

\section{Proof of Lemma~\ref{lm:i-1th_round}}\label{appd:lm:i-1th_round}
	The inference loss of the given $(\psi_{m,1},\ldots,\psi_{m,i-1}, \bar\psi_{m,i})$ is
	\begin{align}
		J(\psi_{m,1},\ldots,\psi_{m,i-1}, \bar\psi_{m,i})
		&= \E\Big[\sum_{j=1}^{i-2} \tilde{\ell}\big((\pi_m, X_j), \wh Y_j\big)\Big] + \nonumber \\
		&\quad \,\, \E\big[ \tilde{\ell}\big((\pi_m, X_{i-1}), \wh Y_{i-1} \big)\big] + \nonumber\\
		&\quad \,\, \E\big[ \tilde{\ell}\big((\pi_m, X_i), \bar\psi_{m,i}(\pi_m, X_i) \big)\big] . \label{eq:pf_i-1_round_lm_2}
	\end{align}
	Since the first expectation in \eqref{eq:pf_i-1_round_lm_2} does not depend on $\psi_{m,i-1}$, it suffices to show that there exists a learned estimator $\bar\psi_{m,i-1}:\Delta\times\sX\rightarrow\wh\sY$, such that 
	\begin{align}
		&\E\big[ \tilde{\ell}\big((\pi_m, X_{i-1}), \bar\psi_{m,i-1}(\pi_m, X_{i-1}) \big)\big] + 
		\E\big[ \tilde{\ell}\big((\pi_m, \bar X_i), \bar\psi_{m,i}(\pi_m, \bar X_i) \big)\big] \nonumber\\
		\le &
		\E\big[ \tilde{\ell}\big((\pi_m, X_{i-1}), \wh Y_{i-1} \big)\big] + 
		\E\big[ \tilde{\ell}\big((\pi_m, X_i), \bar\psi_{m,i}(\pi_m, X_i) \big)\big] , \label{eq:pf_i-1_round_lm_3}
	\end{align}
	where $\bar X_i$ on the left side is the observation in the $i$th round when the Markov offline-learned estimator $\bar\psi_{m,i-1}$ is used in the $(i-1)$th round.
	To get around with the dependence of $X_i$ on $\psi_{m,i-1}$, we write the second expectation on the right side of \eqref{eq:pf_i-1_round_lm_3} as 
	\begin{align}
		\E\big[ \E\big[ \tilde{\ell}\big((\pi_m, X_i), \bar\psi_{m,i}(\pi_m, X_i) \big) \big| \pi_m, X_{i-1}, \wh Y_{i-1}\big] \big]
	\end{align}
	and notice that the conditional expectation $\E\big[ \tilde{\ell}\big((\pi_m, X_i), \bar\psi_i(\pi_m, X_i) \big) \big| \pi_m, X_{i-1}, \wh Y_{i-1}\big]$ does not depend on $\psi_{i-1}$. This is because the conditional distribution of $(\pi_m, X_i)$ given $(\pi_m, X_{i-1}, \wh Y_{i-1})$ is solely determined by the probability transition kernel $P{\raisebox{-2pt}{$\scriptstyle X_i|X_{i-1}, \wh Y_{i-1}$}}$, as shown in the proof of Lemma~\ref{lm:controlled_MC_LDI} stating that $(\pi_m,X_i)_{i=1}^n$ is a controlled Markov chain with $\wh Y^n$ as the control sequence.
	It follows that the right side of \eqref{eq:pf_i-1_round_lm_3} can be written as
	\begin{align}
		& \E\Big[ \tilde{\ell}\big((\pi_m, X_{i-1}), \wh Y_{i-1} \big) + 
		\E\big[ \tilde{\ell}\big((\pi_m, X_i), \bar\psi_{m,i}(\pi_m, X_i) \big) \big| \pi_m, X_{i-1}, \wh Y_{i-1}\big] \Big] \nonumber \\
		=& \E\big[g\big(	\pi_m, X_{i-1}, \wh Y_{i-1} \big)\big] \\
		=& \E\big[g\big(\pi_m, X_{i-1}, \psi_{m,i-1}(Z^m, X^{i-1}, \wh Y^{i-2})\big) \big]
	\end{align}
	for a function $g$ that does not depend on $\psi_{m,i-1}$.
	Since $(\pi_m, X_{i-1})$ is a function of $(Z^m, X^{i-1}, \wh Y^{i-2})$, it follows from Lemma~\ref{lm:gen_Blackwell} that there exists a learned estimator $\bar\psi_{m,i-1}:\Delta\times\sX\rightarrow\wh\sY$, such that 
	\begin{align}
		& \E\big[g\big(\pi_m, X_{i-1}, \psi_{m,i-1}(Z^m, X^{i-1}, \wh Y^{i-2})\big) \big] \\
		\ge & \E\big[g\big(	\pi_m, X_{i-1}, \bar\psi_{m,i-1}(\pi_m, X_{i-1}) \big)\big] \\
		= & \E\Big[ \tilde{\ell}\big((\pi_m, X_{i-1}), \bar\psi_{m,i-1}(\pi_m, X_{i-1}) \big) + \nonumber \\
		&\quad \E\big[ \tilde{\ell}\big((\pi_m, \bar X_i), \bar\psi_{m,i}(\pi_m, \bar X_i) \big) \big| \pi_m, X_{i-1}, \bar\psi_{m,i-1}(\pi_m, X_{i-1})\big] \Big] \\
		= & \E\big[ \tilde{\ell}\big((\pi_m, X_{i-1}), \bar\psi_{m,i-1}(\pi_m, X_{i-1}) \big)\big] + 
		\E\big[ \tilde{\ell}\big((\pi_m, \bar X_i), \bar\psi_{m,i}(\pi_m, \bar X_i) \big)\big] ,
	\end{align}
	which proves \eqref{eq:pf_i-1_round_lm_3} and the claim.

\section{Proof of Theorem~\ref{th:Markov_LDI}}\label{appd:th:Markov_LDI}
	Picking an optimal offline-learned estimation strategy $\psi_m^n$, we can first replace its last estimator by a Markov one that preserves the optimality of the strategy, which is guaranteed by Lemma~\ref{lm:last_round}.
	Then, for $i=n,\ldots,2$, we can repeatedly replace the $(i-1)$th estimator by a Markov one that preserves the optimality of the previous strategy, which is guaranteed by Lemma~\ref{lm:i-1th_round} and the additive structure of the inference loss as in \eqref{eq:def_inf_loss}.
	Finally we obtain an offline-learned estimation strategy consisting of Markov estimators that achieves the same inference loss as the originally picked offline-learned estimation strategy.

\section{Proof of Theorem~\ref{th:opt_dp_LDI}}\label{appd:th:opt_dp_LDI}
	The first claim stating that the offline-learned estimation strategy $(\psi^*_{m,1},\ldots,\psi^*_{m,n})$ achieves the minimum in \eqref{eq:problem_LDI} follows from the equivalence between \eqref{eq:problem_LDI} and the MDP in \eqref{eq:LDT_MDP}, and from the well-known optimality of the solution derived from dynamic programming to MDP.
	
	The second claim can be proved via backward induction.
	Consider an arbitrary Markov offline-learned estimation strategy $\psi_m^n$ with $\psi_{m,i}:\Delta\times\sX\rightarrow\sY$, based on which the learned estimates during inference are made.
	\begin{itemize}[leftmargin=*]
		\item 
		In the final round, for all $\pi\in\Delta$ and $x\in\sX$,
		\begin{align}
			V_{m,n}(\pi, x; \psi_m^n) &= \tilde \ell(\pi, x,\psi_{m,n}(\pi, x)) \label{eq:pf_opt_dp_LDI_1} \\
			&\ge V^*_{m,n}(\pi,x) , \label{eq:pf_opt_dp_LDI_2}
		\end{align}
		where \eqref{eq:pf_opt_dp_LDI_1} is due to the definitions of $V_{m,n}$ in \eqref{eq:def_Vmi} and $\tilde{\ell}$ in \eqref{eq:def_tilde_ell}; and \eqref{eq:pf_opt_dp_LDI_2} is due to the definition of $V^*_{m,n}$ in \eqref{eq:dp_LDI_V}, while the equality holds if $\psi_{m,n}(\pi,x)=\psi^*_{m,n}(\pi,x)$.
		\item
		For $i=n-1,\ldots,1$, suppose \eqref{eq:opt_V_dp_LDI} holds in the $(i+1)$th round. 
		We first show a self-recursive expression of $V_{m,i}(\pi,x; \psi_m^n)$:
		\begin{align}
			V_{m,i}(\pi,x; \psi_m^n) &= \E\Big[ \sum_{j=i}^n \ell(X_j, Y_j, \wh Y_j) \Big| \pi_m = \pi, X_i=x\Big] \\
			&= \E[\ell(X_i,Y_i,\wh Y_i)|\pi_m = \pi, X_i=x] + \E\Big[ \sum_{j=i+1}^n \ell(X_j, Y_j, \wh Y_j) \Big| \pi_m=\pi, X_i=x\Big] \\
			&= \E\big[\E[\ell(X_i,Y_i,\wh Y_i)| \wh Y_i, \pi_m = \pi, X_i=x] \big| \pi_m = \pi, X_i=x\big] + \nonumber \\
			&\quad\,\,  \E\Bigg[\E\Big[ \sum_{j=i+1}^n \ell(X_j, Y_j, \wh Y_j) \Big|X_{i+1}, \pi_m = \pi, X_i=x\Big] \Bigg | \pi_m = \pi,  X_i=x \Bigg] \\
			&= \E\big[\tilde\ell(\pi, x,\wh Y_i) \big|\pi_m = \pi, X_i=x\big] + \nonumber \\
			&\quad\,\,  \E\Bigg[\E\Big[ \sum_{j=i+1}^n \ell(X_j, Y_j, \wh Y_j) \Big|\pi_m = \pi, X_{i+1}\Big] \Bigg | \pi_m = \pi,  X_i=x \Bigg] \label{eq:pf_LDI_V_recur_1} \\
			&= \tilde\ell(\pi, x,\psi_{m,i}(\pi,x)) + \E\big[V_{m,i+1}(\pi, X_{i+1};\psi_m^n) | \pi_m = \pi, X_i=x \big] 
		\end{align}
		where the second term of \eqref{eq:pf_LDI_V_recur_1} follows from the fact that $X_i$ is conditionally independent of $(X_{i+1}^n, Y_{i+1}^n, \wh Y_{i+1}^n)$ given $(\pi_m,X_{i+1})$, which is a consequence of the assumption that the offline-learned estimators are Markov and the specification of the joint distribution of $(Z^m, X^n, Y^n, \wh Y^n)$ in the setup of the offline learning problem, and can be seen from Fig.~\ref{fig:BN_LDI_IL}.
		Then,
		\begin{align}
			V_{m,i}(\pi,x; \psi_m^n) &\ge \tilde\ell(\pi,x,\psi_{m,i}(\pi,x)) + \E\big[V^*_{m,i+1}(\pi,X_{i+1}) | \pi_m = \pi, X_i=x \big] \label{eq:pf_opt_V_dp_LDI_1} \\
			&= \tilde\ell(\pi, x,\psi_{m,i}(\pi,x)) + \E\big[V^*_{m,i+1}(\pi,X_{i+1}) | \pi_m = \pi, X_i=x, \wh Y_i = \psi_{m,i}(\pi,x) \big] \label{eq:pf_opt_V_dp_LDI_2} \\
			&= \tilde\ell(\pi, x,\psi_{m,i}(\pi,x)) + \E\big[V^*_{m,i+1}(\pi,X_{i+1}) | X_i=x, \wh Y_i = \psi_{m,i}(\pi,x) \big] \label{eq:pf_opt_V_dp_LDI_3} \\
			&= Q^*_{m,i}(\pi, x,\psi_{m,i}(\pi,x)) \\
			&\ge V^*_{m,i}(\pi, x)
		\end{align}
		where \eqref{eq:pf_opt_V_dp_LDI_1} follows from the inductive assumption; \eqref{eq:pf_opt_V_dp_LDI_2} follows from the fact that $\wh Y_i$ is determined given $\pi_m = \pi$ and $X_i=x$; \eqref{eq:pf_opt_V_dp_LDI_3} follows from the fact that $X_{i+1}$ is independent of $\pi_m$ given $(X_i,\wh Y_i)$; and the final inequality with the equality condition follow from the definitions of $V^*_{m,i}$ and $\psi^*_{m,i}$ in \eqref{eq:dp_LDI_V} and \eqref{eq:dp_LDI_policy}.
	\end{itemize}
	This proves the second claim.

\section{Proof of Theorem~\ref{th:loss_onLDI}}\label{appd:th:loss_onLDI}
	For each $i=1,\ldots,n$, we have
	\begin{align}
		&\E\big[ {\ell}( X_i, Y_i, \wh Y_i) \big| Z^{i-1}, \wh Y^{i-1}, X_i \big] \nonumber \\
		=& \int_\sY  P_{Y_i|Z^{i-1}, \wh Y^{i-1}, X_i}({\rm d}y) \ell(X_i, y, \wh Y_i) \label{eq:pf_onLDI_loss_1} \\
		=& \int_\sW \int_\sY P_{W|Z^{i-1}, \wh Y^{i-1}, X_i}({\rm d}w) P_{Y_i|Z^{i-1}, \wh Y^{i-1}, X_i,W=w}({\rm d}y) \ell(X_i, y, \wh Y_i) \\
		=& \int_\sW \int_\sY \pi_i({\rm d}w) P_{Y|X,W}({\rm d}y | X_i, w) \ell(X_i, y, \wh Y_i) \label{eq:pf_onLDI_loss_2} \\
		=&\tilde{\ell}(\pi_i, X_i, \wh Y_i) ,
	\end{align}
	where \eqref{eq:pf_onLDI_loss_1} is due to the fact that $X_i$ and $\wh Y_i$ are determined by $(Z^{i-1}, \wh Y^{i-1}, X_i)$; and \eqref{eq:pf_onLDI_loss_2} follows from the fact that $W$ is conditionally independent of $(\wh Y^{i-1},X_i)$ given $Z^{i-1}$ as a consequence of Lemma~\ref{lm:cond_ind_W_onLDI}, and the fact that $Y_i$ is conditionally independent of $(Z^{i-1}, \wh Y^{i-1})$ given $(X_i,W)$.
	With the above equality and the fact that 
	\begin{align}
		\E\Big[ \sum_{i=1}^n \ell(X_i,Y_i,\wh Y_i) \Big] =  \sum_{i=1}^n \E\big[ \E[ {\ell}( X_i, Y_i, \wh Y_i) | Z^{i-1}, \wh Y^{i-1}, X_i ] \big] ,
	\end{align}
	we obtain \eqref{eq:loss_onLDI}.

\section{Proof of Lemma~\ref{lm:controlled_MC_onLDI}}\label{appd:lm:controlled_MC_onLDI}
	We first show that $\pi_{i+1}$ can be determined by $(\pi_i,X_i,Y_i)$.
	To see it, we express $\pi_{i+1}$ as 
	\begin{align}
		P_{W|Z^i} &= {P_{W,Z_i|Z^{i-1}}} / {P_{Z_i|Z^{i-1}}} \\
		&= {P_{W|Z^{i-1}} P_{X_i|W,Z^{i-1}} P_{Y_i|X_i,W,Z^{i-1}} } / {P_{Z_i|Z^{i-1}}} \\
		&= { \pi_{i} P_{X_i|X_{i-1},\wh Y_{i-1}} P_{Y_i|X_i,W} } / {P_{Z_i|Z^{i-1}}}  \label{eq:pf_cMC_onLDI_2} \\
		&= \frac{ \pi_{i} P_{Y_i|X_i,W} } { \int_\sW \pi_{i}({\rm d} w') P_{Y_i|X_i,W=w'} }
	\end{align}
	where \eqref{eq:pf_cMC_onLDI_2} follows from the facts that 1) $\wh Y_{i-1}$ is determined by $Z^{i-1}$, and $X_i$ is conditionally independent of $(W,Z^{i-2},Y_{i-1})$ given $(X_{i-1},\wh Y_{i-1})$; and 2) $Y_i$ is conditionally independent of $Z^{i-1}$ given $(X_i,W)$.
	It follows that $\pi_{i+1}$ can be written as
	\begin{align}\label{eq:pf_cMC_onLDI_3}
		\pi_{i+1} = f(\pi_i,X_i,Y_i)
	\end{align}
	for a function $f$ that maps $\big(\pi_i(\cdot),X_i,Y_i\big)$ to $\pi_{i+1}(\cdot) \propto \pi_{i}(\cdot) P_{Y|X,W}(Y_i|X_i,\cdot)$.
	
	With \eqref{eq:pf_cMC_onLDI_3}, 
	for any online-learned estimation strategy $\psi^n$, any Borel sets $A\subset\Delta$ and $B\subset\sX$, and any realization of $(\pi^i, X^i, \wh Y^i)$, we have 
	\begin{align}
		& \PP\big[  (\pi_{i+1}, X_{i+1}) \in A\times B \big| \pi^i, X^i, \wh Y^i \big ] \nonumber \\
		=& \int_\sY \PP\big[ {\rm d}y_i \big| \pi^i, X^i, \wh Y^i \big ] \PP\big[ (\pi_{i+1}, X_{i+1}) \in A\times B \big| \pi^i, X^i, \wh Y^i, Y_i = y_i  \big ]  \\ 
		=& \int_\sY \PP\big[ {\rm d}y_i \big| \pi^i, X^i, \wh Y^i \big ] \PP\big[ f(\pi_{i}, X_{i},y_i) \in A] \PP\big[X_{i+1}\in B \big| X_i, \wh Y_i \big ] \label{eq:pf_cMC_onLDI_5} \\  
		=& \int_\sY \int_\sW \PP\big[ {\rm d}w \big| \pi^i, X^i, \wh Y^i \big ] \PP\big[ {\rm d}y_i \big| \pi^i, X^i, \wh Y^i, W=w \big ]
		\PP\big[ f(\pi_{i}, X_{i},y_i) \in A] \PP\big[X_{i+1}\in B \big| X_i, \wh Y_i \big ]  \\  
		=&  \int_\sY \int_\sW \pi_i({\rm d}w) P_{Y|X,W}({\rm d}y_i | X_i, w)  \PP[f(\pi_{i}, X_i, y_i) \in A] \PP\big[ X_{i+1} \in B | X_i, \wh Y_i \big] \label{eq:pf_cMC_onLDI_6} , 
	\end{align}
	where \eqref{eq:pf_cMC_onLDI_5} follows from \eqref{eq:pf_cMC_onLDI_3} and the fact that $X_{i+1}$ is conditionally independent of $( Z^{i-1}, Y_i, \wh Y^{i-1})$ given $(X_i,\wh Y_i)$; and \eqref{eq:pf_cMC_onLDI_6} follows from 1) Lemma~\ref{lm:cond_P} and the fact that $W$ is conditionally independent of $(Z^{i-1},X_i,\wh Y^i)$ given $Z^{i-1}$, as a consequence of Lemma~\ref{lm:cond_ind_W_onLDI}, and 2) the fact that $Y_i$ is conditionally independent of $(Z^{i-1}, \wh Y^i)$ given $(X_i,W)$.
	
	This proves the Lemma~\ref{lm:controlled_MC_onLDI}, and we see that the right side of \eqref{eq:pf_controlled_MC_onLDI} only depends on $(\pi_i, X_i,  \wh Y_i)$.

\section{Proof of Lemma~\ref{lm:cond_P}}\label{appd:lm:cond_P}
	Given a probability distribution $p$ on $\sV$, let $\sU_p \deq \{u\in\sU: P_{V|U}(\cdot|u) = p\}$.
	Then, for any Borel sets $ A\in \mathsf V$ and $ B\in\mathsf T$, 
	\begin{align}
		\PP\big[ V \in  A \big| P_{V|U}(\cdot|U) = p, T \in  B \big] 
		&= \frac{ \PP\big[ V \in  A , P_{V|U}(\cdot|U) = p, T \in  B \big] }{ \PP\big[ P_{V|U}(\cdot|U) = p, T \in  B \big] } \\
		&= \frac{ \int_{\sU_p} P_U({\rm d}u) P_{V|U}( A | u) P_{T|U} (B|u)   }{ \int_{\sU_p} P_U({\rm d}u) P_{T|U} (B|u) } \label{eq:pf_lm_cond_P_1} \\
		&= p(A) , \label{eq:pf_lm_cond_P_2}
	\end{align}
	where \eqref{eq:pf_lm_cond_P_1} follows from the definition of $\sU_p$ and the assumption that $T$ and $V$ are conditionally independent given $U$; and \eqref{eq:pf_lm_cond_P_2} follows from the fact that $P_{V|U}(A|u)=p(A)$ for all $u\in\sU_p$.

\section{Proof of Lemma~\ref{lm:last_round_on}}\label{appd:lm:last_round_on}
	The inference loss of $\psi_i^n$ can be written as
	\begin{align}
		J(\psi^{n}) 
		&= \E\Big[\sum_{i=1}^{n-1} \tilde{\ell}\big((\pi_i, X_i), \wh Y_i\big)\Big]
		+ \E\big[ \tilde{\ell}\big((\pi_n, X_n), \psi_{n}(Z^{n-1}, \wh Y^{n-1}, X_n) \big)\big] . \label{eq:pf_last_round_lm_on_2}
	\end{align}
	Since the first expectation in \eqref{eq:pf_last_round_lm_on_2} does not depend on $\psi_{n}$, it suffices to show that there exists a Markov online-learned estimator $\bar\psi_{n}:\Delta\times\sX\rightarrow\wh\sY$, such that 
	\begin{align}
		\E\big[ \tilde{\ell}\big((\pi_n, X_n), \bar\psi_{n}(\pi_n, X_n) \big)\big]  \le
		\E\big[ \tilde{\ell}\big((\pi_n, X_n), \psi_{n}(Z^{n-1}, \wh Y^{n-1}, X_n) \big)\big] .
	\end{align}
	The existence of such an estimator is guaranteed by Lemma~\ref{lm:gen_Blackwell}, as $(\pi_n, X_n)$ is a function of $(Z^{n-1}, \wh Y^{n-1}, X_n)$.

\section{Proof of Lemma~\ref{lm:i-1th_round_on}}\label{appd:lm:i-1th_round_on}
The proof is given in Appendix~\ref{appd:lm:i-1th_round_on}.
	The inference loss of the given $(\psi_{1},\ldots,\psi_{i-1}, \bar\psi_{i})$ is
	\begin{align}
		J(\psi_{1},\ldots,\psi_{i-1}, \bar\psi_{i})
		&= \E\Big[\sum_{j=1}^{i-2} \tilde{\ell}\big((\pi_j, X_j), \wh Y_j\big)\Big] + \nonumber \\
		&\quad \,\, \E\big[ \tilde{\ell}\big((\pi_{i-1}, X_{i-1}), \wh Y_{i-1} \big)\big] + \nonumber\\
		&\quad \,\, \E\big[ \tilde{\ell}\big((\pi_{i}, X_i), \bar\psi_{i}(\pi_i, X_i) \big)\big] . \label{eq:pf_i-1_round_lm_on_2}
	\end{align}
	Since the first expectation in \eqref{eq:pf_i-1_round_lm_on_2} does not depend on $\psi_{i-1}$, it suffices to show that there exists a Markov online-learned estimator $\bar\psi_{i-1}:\Delta\times\sX\rightarrow\wh\sY$, such that 
	\begin{align}
		&\E\big[ \tilde{\ell}\big((\pi_{i-1}, X_{i-1}), \bar\psi_{i-1}(\pi_{i-1}, X_{i-1}) \big)\big] + 
		\E\big[ \tilde{\ell}\big((\pi_i, \bar X_i), \bar\psi_{i}(\pi_i, \bar X_i) \big)\big] \nonumber\\
		\le &
		\E\big[ \tilde{\ell}\big((\pi_{i-1}, X_{i-1}), \wh Y_{i-1} \big)\big] + 
		\E\big[ \tilde{\ell}\big((\pi_i, X_i), \bar\psi_{i}(\pi_i, X_i) \big)\big] , \label{eq:pf_i-1_round_lm_on_3}
	\end{align}
	where $\bar X_i$ on the left side is the observation in the $i$th round when the Markov estimator $\bar\psi_{i-1}$ is used in the $(i-1)$th round.
	To get around with the dependence of $X_i$ on $\psi_{i-1}$, we write the second expectation on the right side of \eqref{eq:pf_i-1_round_lm_on_3} as 
	\begin{align}
		\E\big[ \E\big[ \tilde{\ell}\big((\pi_i, X_i), \bar\psi_{i}(\pi_i, X_i) \big) \big| \pi_{i-1}, X_{i-1}, \wh Y_{i-1}\big] \big]
	\end{align}
	and notice that the conditional expectation $\E\big[ \tilde{\ell}\big((\pi_i, X_i), \bar\psi_i(\pi_i, X_i) \big) \big| \pi_{i-1}, X_{i-1}, \wh Y_{i-1}\big]$ does not depend on $\psi_{i-1}$. This is because the conditional distribution of $(\pi_i, X_i)$ given $(\pi_{i-1}, X_{i-1}, \wh Y_{i-1})$ is solely determined by the probability transition kernels $P_{Y_{i-1}|X_{i-1},W}$ and $P{\raisebox{-2pt}{$\scriptstyle X_i|X_{i-1}, \wh Y_{i-1}$}}$, as shown in the proof of Lemma~\ref{lm:controlled_MC_onLDI} stating that $(\pi_i,X_i)_{i=1}^n$ is a controlled Markov chain driven by $\wh Y^n$.
	It follows that the right side of \eqref{eq:pf_i-1_round_lm_on_3} can be written as
	\begin{align}
		& \E\Big[ \tilde{\ell}\big((\pi_{i-1}, X_{i-1}), \wh Y_{i-1} \big) + 
		\E\big[ \tilde{\ell}\big((\pi_i, X_i), \bar\psi_{i}(\pi_i, X_i) \big) \big| \pi_{i-1}, X_{i-1}, \wh Y_{i-1}\big] \Big] \nonumber \\
		=& \E\big[g\big(	\pi_{i-1}, X_{i-1}, \wh Y_{i-1} \big)\big] \\
		=& \E\big[g\big(\pi_{i-1}, X_{i-1}, \psi_{i-1}(Z^{i-2}, \wh Y^{i-2}, X_{i-1})\big) \big]
	\end{align}
	for a function $g$ that does not depend on $\psi_{i-1}$.
	Since $(\pi_{i-1}, X_{i-1})$ is a function of $(Z^{i-2}, \wh Y^{i-2}, X_{i-1})$, it follows from Lemma~\ref{lm:gen_Blackwell} that there exists a learned estimator $\bar\psi_{i-1}:\Delta\times\sX\rightarrow\wh\sY$, such that 
	\begin{align}
		& \E\big[g\big(\pi_{i-1}, X_{i-1}, \psi_{i-1}(Z^{i-2}, \wh Y^{i-2}, X_{i-1})\big) \big] \\
		\ge & \E\big[g\big(	\pi_{i-1}, X_{i-1}, \bar\psi_{i-1}(\pi_{i-1}, X_{i-1}) \big)\big] \\
		= & \E\Big[ \tilde{\ell}\big((\pi_{i-1}, X_{i-1}), \bar\psi_{i-1}(\pi_{i-1}, X_{i-1}) \big) + \nonumber \\
		&\quad \E\big[ \tilde{\ell}\big(( \pi_i, \bar X_i), \bar\psi_{i}( \pi_i, \bar X_i) \big) \big| \pi_{i-1}, X_{i-1}, \bar\psi_{i-1}(\pi_{i-1}, X_{i-1})\big] \Big] \\
		= & \E\big[ \tilde{\ell}\big((\pi_{i-1}, X_{i-1}), \bar\psi_{i-1}(\pi_{i-1}, X_{i-1}) \big)\big] + 
		\E\big[ \tilde{\ell}\big(( \pi_i, \bar X_i), \bar\psi_{i}( \pi_i, \bar X_i) \big)\big] ,
	\end{align}
	which proves \eqref{eq:pf_i-1_round_lm_on_3} and the claim.

\section{Proof of Theorem~\ref{th:Markov_onLDI}}\label{appd:th:Markov_onLDI}
	Picking an optimal online-learned estimation strategy $\psi^n$, we can first replace its last estimator by a Markov one that preserves the optimality of the strategy, which is guaranteed by Lemma~\ref{lm:last_round_on}.
	Then, for $i=n,\ldots,2$, we can repeatedly replace the $(i-1)$th estimator by a Markov one that preserves the optimality of the previous strategy, which is guaranteed by Lemma~\ref{lm:i-1th_round_on} and the additive structure of the inference loss as in \eqref{eq:def_inf_loss_on}.
	Finally we obtain an online-learned estimation strategy consisting of Markov online-learned estimators that achieves the same inference loss as the originally picked online-learned estimation strategy.

\section{Proof of Theorem~\ref{th:opt_dp_onLDI}}\label{appd:th:opt_dp_onLDI}
	The first claim stating that the online-learned estimation strategy $(\psi^*_{1},\ldots,\psi^*_{n})$ achieves the minimum in \eqref{eq:problem_onLDI} follows from the equivalence between \eqref{eq:problem_onLDI} and the MDP in \eqref{eq:onLDT_MDP}, and from the well-known optimality of the solution derived from dynamic programming to MDP.
	
	The second claim can be proved via backward induction.
	Consider an arbitrary Markov online-learned estimation strategy $\psi^n$ with $\psi_{i}:\Delta\times\sX\rightarrow\sY$, based on which the learned estimates are made.
	For any pair $(i,j)$ such that $1\le i\le j\le n$,
	\begin{align}
		&\E\big[ \ell(X_j, Y_j, \wh Y_j) \big| \pi_i, X_i \big]  \nonumber \\
		=& \E\big[ \E[ \ell(X_j, Y_j, \wh Y_j) | \pi_j, X_j, \pi_i, X_i ] \big| \pi_i, X_i \big] \\
		=& \E\Big[ \int_\sW P({\rm d}w | \pi_j, X_j, \pi_i, X_i) \int_\sY P({\rm d}y_j | \pi_j, X_j, \pi_i, X_i, W=w) \ell(X_j, y_j, \wh Y_j)  \Big| \pi_i, X_i \Big] \label{eq:pf_inf_loss_on_1} \\
		=& \E\Big[ \int_\sW \int_\sY \pi_j({\rm d}w ) P_{Y|X,W}({\rm d}y_j | X_j, w) \ell(X_j, y_j, \wh Y_j)  \Big| \pi_i, X_i \Big] \label{eq:pf_inf_loss_on_2} \\
		=&\E\big[ \tilde\ell(\pi_j, X_j, \wh Y_j) \big| \pi_i, X_i \big] \label{eq:pf_inf_loss_on_3}
	\end{align}
	where \eqref{eq:pf_inf_loss_on_1} follows from the fact that $\wh Y_j$ is determined by $(\pi_j,X_j)$; \eqref{eq:pf_inf_loss_on_2} follows from 1) Lemma~\ref{lm:cond_P} and the fact that $W$ is conditionally independent of $(Z^{i-1}, X_i, X_j)$ given $Z^{j-1}$, and 2) $Y_j$ is conditionally independent of $Z^{j-1}$ given $(X_j,W)$; and \eqref{eq:pf_inf_loss_on_3} follows from the definition of $\tilde{\ell}$ in  \eqref{eq:def_tilde_ell}.
	With the above identity, the loss-to-go defined in \eqref{eq:def_Vi_on} can be rewritten as 
	\begin{align}
		V_{i}(\pi, x; \psi^n) = \E\Big[ \sum_{j=i}^n \tilde\ell(\pi_j, X_j, \wh Y_j) \Big| \pi_i = \pi, X_i=x\Big] , \quad i=1,\ldots,n .
	\end{align}
	Now we can proceed with proving the second claim via backward induction.
	\begin{itemize}[leftmargin=*]
		\item 
		In the final round, for all $\pi\in\Delta$ and $x\in\sX$,
		\begin{align}
			V_{n}(\pi, x; \psi^n) &= \tilde \ell(\pi, x,\psi_{n}(\pi, x)) \label{eq:pf_opt_dp_onLDI_1} \\
			&\ge V^*_{n}(\pi,x) , \label{eq:pf_opt_dp_onLDI_2}
		\end{align}
		where \eqref{eq:pf_opt_dp_onLDI_1} is due to \eqref{eq:pf_inf_loss_on_3} with $i=j=n$; and \eqref{eq:pf_opt_dp_onLDI_2} is due to the definition of $V^*_{n}$ in \eqref{eq:dp_onLDI_V}, while the equality holds if $\psi_{n}(\pi,x)=\psi^*_{n}(\pi,x)$.
		\item
		For $i=n-1,\ldots,1$, suppose \eqref{eq:opt_V_dp_onLDI} holds in the $(i+1)$th round. 
		We first show a self-recursive expression of $V_{i}(\pi,x; \psi^n)$:
		\begin{align}
			&\quad\, V_{i}(\pi,x; \psi^n) \nonumber \\
			&= \E\Big[ \sum_{j=i}^n \tilde\ell(\pi_j, X_j, \wh Y_j) \Big| \pi_i = \pi, X_i=x\Big] \\
			&= \E[\tilde\ell(\pi_i,X_i,\wh Y_i)|\pi_i = \pi, X_i=x] + \E\Big[ \sum_{j=i+1}^n \tilde\ell(\pi_j, X_j, \wh Y_j) \Big| \pi_i=\pi, X_i=x\Big] \\
			&= \tilde\ell(\pi, x,\psi_{i}(\pi,x))  +  
			\E\Bigg[\E\Big[ \sum_{j=i+1}^n \tilde\ell(\pi_j, X_j, \wh Y_j) \Big|\pi_{i+1},X_{i+1}, \pi_i = \pi, X_i=x\Big] \Bigg | \pi_i = \pi,  X_i=x \Bigg] \\
			&= \tilde\ell(\pi, x,\psi_{i}(\pi,x)) +  
			\E\Bigg[\E\Big[ \sum_{j=i+1}^n \tilde\ell(\pi_j, X_j, \wh Y_j) \Big|\pi_{i+1}, X_{i+1}\Big] \Bigg | \pi_i = \pi,  X_i=x \Bigg] \label{eq:pf_onLDI_V_recur_1} \\
			&= \tilde\ell(\pi, x,\psi_{i}(\pi,x)) + \E\big[V_{i+1}(\pi_{i+1}, X_{i+1};\psi^n) | \pi_i = \pi, X_i=x \big] 
		\end{align}
		where the second term of \eqref{eq:pf_onLDI_V_recur_1} follows from the fact that $\wh Y_{i+1}$ is determined by $(\pi_{i+1},X_{i+1})$, and the fact that $(\pi_j,X_j)_{j=i+1}^n$ is conditionally independent of $(\pi_i,X_i)$ given $(\pi_{i+1}, X_{i+1}, \wh Y_{i+1})$ as guaranteed by Lemma~\ref{lm:controlled_MC_onLDI}.
		Then,
		\begin{align}
			V_{i}(\pi,x; \psi^n) &\ge \tilde\ell(\pi,x,\psi_{i}(\pi,x)) + \E\big[V^*_{i+1}(\pi_{i+1}, X_{i+1}) | \pi_i = \pi, X_i=x \big] \label{eq:pf_opt_V_dp_onLDI_1} \\
			&= \tilde\ell(\pi, x,\psi_{i}(\pi,x)) + \E\big[V^*_{i+1}(\pi_{i+1}, X_{i+1}) | \pi_i = \pi, X_i=x, \wh Y_i = \psi_{i}(\pi,x) \big] \label{eq:pf_opt_V_dp_onLDI_2} \\
			&= Q^*_{i}(\pi, x,\psi_{i}(\pi,x)) \label{eq:pf_opt_V_dp_onLDI_3} \\
			&\ge V^*_{i}(\pi, x)
		\end{align}
		where \eqref{eq:pf_opt_V_dp_onLDI_1} follows from the inductive assumption; \eqref{eq:pf_opt_V_dp_onLDI_2} follows from the fact that $\wh Y_i$ is determined given $(\pi_i, X_i)$; \eqref{eq:pf_opt_V_dp_onLDI_3} follows from the definition of $Q^*_i$ in \eqref{eq:dp_onLDI_Q}; and the final inequality with the equality condition follow from the definitions of $V^*_{i}$ and $\psi^*_{i}$ in \eqref{eq:dp_onLDI_V} and \eqref{eq:dp_onLDI_policy}.
	\end{itemize}
	This proves the second claim.

\section*{Acknowledgement}
The authors would like to thank Prof.\ Maxim Raginsky for the encouragement of looking into dynamic aspects of statistical problems, and Prof.\ Lav Varshney for helpful discussions on this work.

\bibliography{DI_bib}

\blfootnote{author email: xuaolin@gmail.com, guanpeng333@gmail.com}

\end{document}